\documentclass[12pt]{iopart}
\pdfoutput=1

\usepackage{iopams}  
\usepackage{amssymb}
\usepackage{hyperref}
\usepackage{lineno}
\usepackage{afterpage}
\usepackage[ruled]{algorithm2e}
\usepackage{chemfig}
\usepackage{chemscheme}
\usepackage{wasysym}

\newcommand{\bx}{\mathbf{x}}
\newcommand{\by}{\mathbf{y}}

\newcommand{\bff}{\mathbf{f}}
\newcommand{\bg}{\mathbf{g}}

\newcommand{\bu}{\mathbf{u}}

\newcommand{\bss}{\mathbf{s}}

\newcommand{\bI}{\mathbf{I}}
\newcommand{\calN}{\mathcal{N}}
\newcommand{\calQ}{\mathcal{Q}}

\newcommand{\calF}{\mathcal{F}}
\newcommand{\calH}{\mathcal{H}}
\newcommand{\calL}{\mathcal{L}}

\newcommand{\trace}[1]{\operatorname{Tr}\left[#1\right]}

\newcommand{\operatorname}[1]{\mbox{#1}}
\newcommand{\E}[2]{\mathbb{E}_{#1}\left[#2\right]}           % Expecation operator
\newcommand{\Eq}[1]{\mathbb{E}_q\left[#1\right]}

\newcommand{\KL}[2]{\operatorname{KL}\left[#1\parallel #2\right]}                % Kullback-Leibler

\newcommand{\R}{\mathbb{R}}

\newcommand{\nothree}[1]{\left[\operatorname{NO}_3^-\right]}

\newcommand{\eqref}[1]{Eq.~(\ref{eqn:#1})}
\newcommand{\aref}[1]{Algorithm~\ref{alg:#1}}
\begin{document}

\title[]{Variational Reformulation of Bayesian Inverse Problems}

\author{Panagiotis Tsilifis$^1$, Ilias Bilionis$^2$\footnote{Corresponding author.},
Ioannis Katsounaros$^{3a,3b,3c}$, Nicholas Zabaras$^6$}

\address{$^1$Department of Mathematics, University of Southern
  California, Los Angeles, CA 90089-2532, USA}
\address{$^2$School of Mechanical Engineering, Purdue University,
585 Purdue Mall,
West Lafayette, IN 47906-2088, USA}
\address{$^{3a}$Department of Chemistry, University of Illinois at Urbana-Champaign,
S. Mathews Ave., Urbana,  IL 61801, USA}
\address{$^{3b}$Materials Science Division, Argonne National Laboratory,
9700 S. Cass Ave., Lemont, IL 60439, USA}
\address{$^{3c}$Leiden Institute of Chemistry, Leiden University, Einsteinweg 55, P.O.
Box 9502, 2300 RA Leiden, The Netherlands}
\address{$^4$Warwick Centre for Predictive Modelling, The University of Warwick, Coventry, CV4 7AL, UK}

\ead{\{tsilifis@usc.edu, ibilion@purdue.edu, katsounaros@anl.gov, nzabaras@gmail.com\}}

\begin{abstract}
The classical approach to inverse problems is based on the optimization of a
misfit function.
Despite its computational appeal, such an approach suffers from many
shortcomings, e.g., non-uniqueness of solutions, modeling prior knowledge, etc.
The Bayesian formalism to inverse problems avoids most of the difficulties
encountered by the optimization approach, albeit at an increased computational
cost.
In this work, we use information theoretic arguments to cast the
Bayesian inference problem in terms of an optimization problem.
The resulting scheme combines the theoretical soundness of fully Bayesian
inference with the computational efficiency of a simple optimization.
\end{abstract}

\submitto{\IP}

\section{Introduction}
\label{sec:intro}

As we are approaching the era of exascale computing, we
encounter more and more complex physical models.
These complex models have many unknown parameters that need to
be inferred from experimental measurements.
That is, inverse problems are becoming an integral part of every scientific
discipline that attempts to combine computational models with data.
These include climate modeling~\cite{bilionisCrop2014},
numerical weather prediction~\cite{citeulike:8803391,ionel2009},
subsurface hydrology and geology~\cite{mary2006effective},
and many more.

It can be argued that the ``right" way to pose an inverse problem is
to follow the Bayesian
formalism~\cite{tarantola2005inverse,jaynes2003probability}.
It is the ``right" way because it deals with three well-known difficulties of
inverse problems: non-uniqueness issues, modeling prior
knowledge, and estimating experimental noise.
The Bayesian solution of an inverse problem is summarized neatly by the
\emph{posterior} probability density of the quantity of interest.
In turn, the posterior can only be explored numerically by Monte Carlo (MC)
methods, the most successful of which is Markov Chain Monte Carlo (MCMC)~\cite{annealing2,hastings1970montecarlo}.
Because of the need to repeatedly evaluate the underlying physical
model, MCMC is computationally impractical for all but the simplest cases.
Therefore, we need methods that approximate the posterior in a computationally efficient
manner.

One way to come up with a computationally attractive approximation is to replace the physical model with
a cheap-to-evaluate surrogate~\cite{Kennedy_2001d,citeulike:12630613}.
In this way, MCMC becomes feasible again.
However, there is no free lunch: firstly, things become complicated when the
surrogate is inaccurate, and, secondly, constructing accurate
surrogates is exponentially hard as the number of parameters increase~\cite{bilionis2014solution}.
Because of these difficulties, in this work, we attempt to develop a surrogate-free
methodology.

Perhaps the simplest idea is to approximate the posterior with a
delta function centered about its maximum.
The maximum of the posterior is known as the maximum a posteriori (MAP)
estimate of the parameters.
The MAP approach is  justified if the posterior is sharply picked around a
unique maximum.
It requires the solution of an optimization problem.
The objective function of this optimization has two parts: a misfit and a
regularization part that arise from the likelihood and the prior, respectively.
The MAP estimate is commonly used in numerical weather prediction problems~\cite{ionel2009}
as well as in seismic inversion~\cite{fichtner2010full}.

The Laplace approximation represents the posterior by a Gaussian
density with a mean given by the MAP and a covariance matrix given by the
negative inverse Hessian of the logarithm of the posterior.
The Laplace approximation is justified when the posterior has a unique maximum
and is shaped, more or less, like a Gaussian around it.
As before, the identification of the MAP requires the solution of an optimization
problem.
The required Hessian information may be estimated numerically either by
automatic differentiation methods~\cite{griewank2008evaluating} or by developing the adjoint
equations of the underlying physical model~\cite{Plessix2006}.

Variational inference (VI)~\cite{Fox2011,Peng2014} is a class of techniques in Bayesian
statistics targeted toward the construction of approximate posteriors.
One proceeds by posing a variational problem whose solution over a family of
probability densities approximates the posterior.
VI techniques have been proved quite effective for a wide class of inference
problems.
However, in their archetypal form, they are not directly applicable to inverse
problems.
This is due to the, typically, non-analytic nature of the underlying physical
 models.
Fortunately, the recent developments in non-parametric VI by~\cite{gershman2012}
can come to rescue.
This is the approach we follow in this work.
In non-parametric VI, the family of probability densities that serve as
candidate posteriors is the family of mixtures of Gaussians with a
fixed number of components~\cite{McLachlan2004}.
Since a mixture of Gaussians with an adequate number of components can
represent any probability density, this approximating
family is sufficiently large.
The approximate posterior is constructed by minimizing the information loss
between the true posterior and the approximate one over the candidate family.
This is achieved by solving a series of optimization problems~\cite{Peng2014}.

The outline of the paper is as follows.
We start \sref{sec:metho} with a generic discussion of the Bayesian formulation of
inverse problems.
In \sref{sec:vi} we present the core ideas of VI and in
\sref{sec:approximations} we show how one can develop approximation schemes.
We validate the proposed methodology numerically by solving two inverse problems:
the estimation of the kinetic parameters of a catalysis system (\sref{sec:react})
and the identification of the source of contamination based on scarce
concentration measurements (\sref{sec:contamination}).
Finally, in \ref{ap:F_grad} we provide all the technical details that are
required for the implementation of the proposed methodology. 

\section{Methodology}
\label{sec:metho}

A forward model associated with a physical phenomenon can be thought of as a
function $\bff:\R^{d_{\xi}}\rightarrow \R^{d_y}$, that connects some unknown parameters
$\bxi\in\R^{d_{\xi}}$ to some observed quantities $\by\in\R^{d_y}$.
This connection is defined indirectly via a
\emph{likelihood} function:
\begin{equation}
    p(\by|\bxi, \btheta) = p(\by | \bff(\bxi), \btheta),
    \label{eqn:likelihood}
\end{equation}
where $\btheta\in\R^{d_{\theta}}$ denotes the parameters that control the measurement noise and/or
the model discrepancy.
Notice how in \eqref{likelihood} the observations, $\by$, are explicitly
connected to the parameters, $\bxi$, through the model, $\bff(\bxi)$.
A typical example of a likelihood function is the \emph{isotropic Gaussian likelihood}:
\begin{equation}
    p(\by | \bxi, \theta) = \calN\left(\by|\bff(\bxi), e^{2\theta} \bI \right),
    \label{eqn:iso_gauss_like}
\end{equation}
where $\theta$ is a real number, and $\bI$ is the unit matrix with
the same dimensions as $\by$.
The exponential of the parameter,
\begin{equation}
\label{eqn:iso_gauss_like_noise}
\sigma = e^{\theta},
\end{equation}
can be thought of as the standard deviation of the measurement noise.
The usual parameterization of the isotropic Gaussian likelihood uses $\sigma$
instead of $\theta$.
We do not follow the usual approach because in our numerical examples,
it is preferable to work with real rather than positive numbers.

Both $\bxi$ and $\btheta$ are characterized by \emph{prior} probability densities,
$p(\bxi)$ and $p(\btheta)$, respectively.
These describe our state of knowledge, prior to observing $\by$.
As soon as $\by$ is observed, our updated state of knowledge is captured by the
\emph{posterior} distribution:
\begin{equation}
    \label{eqn:post}
    p(\bxi, \btheta | \by) = \frac{p(\by|\bxi, \btheta) p(\bxi) p(\btheta)}{p(\by)},
\end{equation}
where the normalization constant $p(\by)$,
\begin{equation}
    \label{eqn:evidence}
    p(\by) = \int p(\by | \bxi, \btheta) p(\bxi) p(\btheta) d\bxi d\btheta,
\end{equation}
is known as the \emph{evidence}.
\eqref{post} is the Bayesian solution to the inverse problem.
Notice that it is a probability density over the joint space of $\bxi$ and $\btheta$.
This is to be contrasted with the classical approaches to inverse problems whose
solutions result in point estimates of the unknown variables.
The mass of this probability density corresponds to our inability to fully
resolve the values of $\bxi$ and $\btheta$ due to insufficient experimental
information.

Notice that by writing $\bomega = (\bxi, \btheta)\in\R^d$,$d = d_{\xi} + d_{\omega}$,
$p(\bomega) = p(\bxi)p(\btheta)$,
and $p(\by|\bomega) = p(\by|\bxi, \btheta)$, we may simplify the notation of \eqref{post}
to
\begin{equation}
    \label{eqn:post_simple}
    p(\bomega|\by) = \frac{p(\by | \bomega)p(\bomega)}{p(\by)}.
\end{equation}
This is the notation we follow through out this work.
The goal of the rest of the paper is to construct an algorithmic framework for
the efficient approximation of the posterior of \eqref{post_simple}.

\subsection{Variational inference}
\label{sec:vi}
Consider a family $\calQ$ of probability densities over $\bomega$.
Our objective is to choose a $q(\bomega) \in \calQ$ that is as ``close" as
possible to the posterior $p(\bomega | \by)$ of \eqref{post_simple}.
This ``closeness" is measured by the Kullback-Leibler (KL) divergence
\cite{citeulike:1649749},
\begin{equation}
\label{eqn:kldiv}
\KL{q(\bomega)}{p(\bomega|\by)} = \Eq{\log \frac{q(\bomega)}{p(\bomega|\by)}},
\end{equation}
where $\Eq{\cdot}$ denotes the expectation with respect to $q(\bomega)$.
Intuitively, the KL divergence can be thought of as the \emph{information loss}
we experience when we approximate the posterior $p(\bomega|\by)$ with the
probability density $q(\bomega)$.
It is easy to show that
\begin{equation}
\label{eqn:kl_inequality}
\KL{q(\bomega)}{p(\bomega|\by)} \ge 0,
\end{equation}
and that the
equality holds if and only if $q(\bomega) = p(\bomega|\by)$.
Therefore, if the posterior is in $\calQ$, then minimizing
\eqref{kldiv} over $q(\omega)\in\calQ$ will give an exact answer.
For an arbitrary choice of $\calQ$, we postulate that minimizing \eqref{kldiv}
yields a good approximation to the posterior.

Unfortunately, calculation of \eqref{kldiv} requires knowledge of the posterior.
This means that \eqref{kldiv} cannot be used directly in an optimization scheme.
In order to proceed, we need an objective function that does not depend explicitly on the posterior.
To this end, notice that the evidence may be expressed as:
\begin{equation}
    \label{eqn:evidence_exp}
\log p(\by) = \calF\left[q\right] + \KL{q(\bomega)}{p(\bomega|\by)},
\end{equation}
where
\begin{equation}
\label{eqn:freeenergy}
\calF\left[q\right] = \Eq{\log\frac{p(\by,\bomega)}{q(\bomega)}} = \calH\left[q\right] + \Eq{\log p(\by, \bomega)},
\end{equation}
with
\begin{equation}
\label{eqn:joint}
p(\by, \bomega) = p(\by | \bomega) p(\bomega)
\end{equation}
being the joint probability density of $\by$ and $\bomega$, and
\begin{equation}
        \label{eqn:entropy}
        \calH[q] = -\Eq{\log q(\bomega)}
\end{equation}
being the \emph{entropy} of $q(\bomega)$.
Since the KL divergence is non-negative (\eqref{kl_inequality}),
we have from \eqref{evidence_exp} that
\begin{equation}
\label{eqn:F_inequality}
\calF[q] \le \log p(\by).
\end{equation}
The functional $\calF[q]$ is generally known as the
\emph{evidence lower bound} (ELBO).

We see, that maximizing \eqref{freeenergy} is equivalent to minimizing
\eqref{kldiv}.
In addition, \eqref{freeenergy} does not depend on the posterior in an explicit
manner.
This brings us to the variational principle used through out this work:
The ``best" approximation to the posterior of \eqref{post_simple} over the
family of probability densities $\calQ$ is the solution to the following
optimization problem:
\begin{equation}
    \label{eqn:opt_full}
    q^*(\bomega) = \arg\max_{q}\calF[q].
\end{equation}

\subsection{Developing approximation schemes}
\label{sec:approximations}
The main difficulty involved in the solution \eqref{opt_full} is the evaluation
of expectations over $q(\bomega)$.
In principle, one can approximate these expectations via a Monte Carlo
procedure and, then, use a variant of the Robbins-Monro algorithm
\cite{robbins1951}.
Such an approach yields a stochastic optimization scheme in the spirit of
\cite{bilionis2013a}, and~\cite{bilionis2012}.
Whether or not such a scheme is more efficient than MCMC sampling of the posterior
is beyond the scope of this work.
Here, we follow the approach outlined by~\cite{gershman2012}.
In particular, we will derive analytical approximations of $\calF[q]$ for the
special case in which $\calQ$ is the family of Gaussian mixtures.

The family of Gaussian mixtures with $L$ components is the family $\calQ_L$ of
probability densities of the form:
\begin{equation}
    \label{eqn:mixture}
q(\bomega) = \sum_{i=1}^Lw_i\calN(\bomega| \bmu_i, \bSigma_i)
\end{equation}
where $w_i$, $\bmu_i$ and $\bSigma_i$ are the responsibility, mean,
and covariance matrix, respectively, of the $i$-th component of the mixture.
The responsibilities $w_i$ are non-negative and they sum to one while the covariance
matrices $\bSigma_i$ are positive definite.
When we work with $\calQ_L$, the generic variational problem of
\eqref{post_simple} is equivalent to optimization with respect to all the
$w_i, \bmu_i$ and $\bSigma_i$.
In what follows, we replace the ELBO, $\calF[q]$ of \eqref{freeenergy}, with
a series of analytic approximations
that exploit the properties of $\calQ_L$, and, finally, we derive a three-step
optimization scheme that yields a local maximum of the approximate ELBO.

We start with an approximation to the entropy $\calH[q]$ (see \eqref{entropy})
of a Gaussian mixture \eqref{mixture}.
There are basically two kinds of approximations that may be derived:
1) using Jensen's inequality yields a lower bound to $\calH[q]$ built out of
convolutions of Gaussians
(see~\cite{gershman2012} and~\cite{ies_2008_huber_mfi_entropy}), and 2)
employing a Taylor expansion of $\log q(\bomega)$ about each $\bmu_i$ and
evaluating the expectation over $\calN(\bomega|\bmu_i, \bSigma_i)$
(see~\cite{ies_2008_huber_mfi_entropy}).
We have experimented with both approximations to the entropy without observing
any significant differences in the numerical results.
Therefore, we opt for the former one since it has a very simple
analytical form.
An application of Jensen's inequality followed by well-known results about the
convolution of two Gaussians yields
\begin{equation}
    \label{eqn:entropy_0}
    \calH[q] \ge \calH_0[q],
\end{equation}
where
\begin{equation}
        \label{eqn:entropy_1}
        \calH_0[q] = -\sum_{i=1}^L w_i \ln q_i,
\end{equation}
with
\begin{equation}
q_i = \sum_{j=1}^L w_j \calN(\bmu_i | \bmu_j, \bSigma_i + \bSigma_j).
\end{equation}
The idea is to simply replace $\calH[q]$ in \eqref{freeenergy} with $\calH_0[q]$ of
\eqref{entropy_1}.
This results in a lower bound to the ELBO $\calF[q]$.

Now, we turn our attention to the second term of \eqref{freeenergy}.
For convenience, let us write it as:
\begin{equation}
        \label{eqn:exp_term}
        \calL[q] = \Eq{\log p(\by, \bomega)}.
\end{equation}
Notice that $\calL[q]$ may be expanded as:
\begin{equation}
\label{eqn:exp_term_expand}
\calL[q] = \sum_{i=1}^L w_i \E{\calN(\bomega|\bmu_i, \bSigma_i)}{\log p(\by, \bomega)},
\end{equation}
and that each expectation term can be approximated by taking the Taylor expansion
of $\log p(\by, \bomega)$ about $\bomega = \bmu_i$:
\begin{equation}
\label{eqn:log_joint_taylor}
\begin{array}{ccc}
\log p(\by, \bomega) &\approx& \log p(\by, \bomega = \bmu_i) + \nabla_{\bomega}
 \log p(\by, \bomega = \bmu_i)\left(\bomega - \bmu_i \right)\\
&& + \frac{1}{2} (\bomega - \bmu_i)^T \nabla^2_{\bomega}\log p(\by, \bomega = \bmu_i) (\bomega - \bmu_i),
\end{array}
\end{equation}
where $\nabla_{\bomega}$ and $\nabla_{\bomega}^2$ stand for the Jacobian and the
Hessian with respect to $\bomega$, respectively.
Combining \eqref{exp_term_expand} with \eqref{log_joint_taylor}, we get the zero
and second order Taylor approximation to $\calL[q]$ of \eqref{exp_term},
\begin{equation}
        \label{eqn:exp_term_0}
        \calL_0[q] = \sum_{i=1}^L w_i \log p(\by, \bomega = \bmu_i),
\end{equation}
and
\begin{equation}
        \label{eqn:exp_term_1}
        \calL_2[q] = \calL_0[q] + \frac{1}{2} \sum_{i=1}^L w_i \trace{\bSigma_i \nabla_{\bomega}^2 \log p(\by, \bomega = \bmu_i)},
\end{equation}
respectively.

Combining \eqref{entropy_1} with \eqref{exp_term_0} or \eqref{exp_term_1} we get
an approximation to \eqref{freeenergy}.
In particular, we define:
\begin{equation}
    \label{eqn:elbo}
        \calF_{a}[q] = \calH_0[q] + \calL_a[q],
\end{equation}
where $a=1$, or $2$, selects the approximation to \eqref{exp_term}.
From this point on, our goal is to derive an algorithm that converges to a
local maximum of $\calF_{2}[q]$.

Notice that $\calF_{2}[q]$ requires the Hessian of $\log p(\by, \bomega)$
at $\bomega = \bmu_i$.
Therefore, optimizing it with respect to $\bmu_i$ would require third derivatives
of $\log p(\by, \bomega)$.
This, in turn, implies the availability of third derivatives of the forward model
$\bff(\bxi)$.
Getting third derivatives of the forward model is impractical in almost all cases.
In contrast, optimization of $\calF_{0}[q]$ with respect to $\bmu_i$ requires
only first derivatives of $\log p(\by, \bomega)$, i.e., only first derivatives of the
forward model $\bff(\bxi)$. In many inverse problems of interest, derivatives
can be obtained at a minimum cost by making use of adjoint techniques
(e.g.,~\cite{Fichtner2011}).
Therefore, optimization of $\calF_{00}[q]$ with respect to $\bmu_i$ is
computationally feasible.

The situation for $\bSigma_i$ is quite different. Firstly, notice that
$\calH_0[q]$ increases logarithmically without bounds as a function of
$|\bSigma_i|$ and that the $\calL_0[q]$ does not depend on $\bSigma_i$ at all.
We see that the lowest approximation that carries information about $\bSigma_i$
is $\calF_{2}[q]$. Looking at \eqref{exp_term_1} we observe that this information
is conveyed via the Hessian of $\log p(\by, \bomega)$. This, in turn, requires
the Hessian of the forward model $\bff(\bxi)$.
The latter is a non-trivial task which is, however, feasible.
In addition, notice that if $\bSigma_i$ is restricted to be diagonal, then only
the diagonal part of the Hessian of $\log p(\by, \bomega)$ is required,
thus, significantly reducing
the memory requirements.
The computation of $F_2[q]$ as well as of its gradients with respect to
$w_i, \bmu_i$, and $\bSigma_i$ is discussed in \ref{ap:F_grad}.

Taking the above into account, we propose an optimization scheme that alternates
between optimizing $\{\bmu_i\}_{i=1}^L$, $\{\bSigma_i\}_{i=1}^L$, and
$\{w_i\}_{i=1}^L$.
The algorithm is summarized in \aref{vi}.
However, in order to avoid the use of third derivatives of the forward model
we follow~\cite{gershman2012} in using $\calF_0[q]$ as the objective function
when optimizing for $\{\bmu_i\}$.
Furthermore, we restrict our attention to diagonal covariance matrices,
\begin{equation}
\label{eqn:diag_covar}
\bSigma_i = \operatorname{diag}\left(\sigma_{i1}^2,\dots,\sigma_{id}^2\right),
\end{equation}
since we do not want to deal with the issue of enforcing positive definiteness
of the $\bSigma_i$'s during their optimization.
We use the L-BFGS-B algorithm of~\cite{Byrd1995}, which can perform bound constrained optimization.
The bounds we use are problem-specific and are discussed in \sref{sec:examples}.
In all numerical examples, we use the same convergence tolerance
$\epsilon=10^{-2}$.

\begin{algorithm}[hbt]
\label{alg:vi}
\SetKwInOut{Input}{Input}\SetKwInOut{Output}{Output}
\SetKwInOut{Initialize}{Initialize}
\DontPrintSemicolon
\Input{Data $\by$, number of components $L$.}
\Initialize{$w_i=1/N, \bSigma_i=I$, and $\bmu_i$ randomly, for $i=1,\dots,L$.}
\Repeat{ change in $\calF_{2}[q]$ is less than a tolerance $\epsilon$}{
  \For{$i = 1$ to $L$}{
    $\{\bmu_i\} \leftarrow \arg\max_{\{\bmu_i\}} \calF_{0}[q]$\\
    $\{w_i\} \leftarrow \arg\max_{\{w_i\}} \calF_{2}[q]$\\
    $\{\bSigma_i\} \leftarrow \arg\max_{\left\{\bSigma_i=\operatorname{diag}\left( \sigma_{i1}^2,\dots,\sigma_{id}^2\right)\right\}} \calF_{2}[q]$
    }
}
\caption{Variational Inference}
\end{algorithm} 

\section{Examples}
\label{sec:examples}

In this section we present two numerical examples: 1) The problem of estimating
rate constants in a catalysis system (\sref{sec:react}), and 2) the problem of
identifying the source of contamination in a two dimensional domain
(\sref{sec:contamination}).
In both examples, we compare the approximate posterior to a MCMC~\cite{hastings1970montecarlo} estimate.
We used the Metropolis-Adjusted-Langevin-Algorithm (MALA)~\cite{Atchade2006},
since it can use the derivatives of the forward models to accelerate
convergence.
In all examples, the step size of the MALA proposal was $dt = 0.1$, the first
$1,000$ steps were burned, and we observed the chain every $100$ steps for a
total of $100,000$ steps.
We implented our methodology is Python.
The code is freely available at \url{https://github.com/ebilionis/variational-reformulation-of-inverse-problems}.

\subsection{Reaction kinetic model}
\label{sec:react}

\begin{figure}[!htb]
\label{fig:reduction}
\begin{center}
\includegraphics[width=120mm]{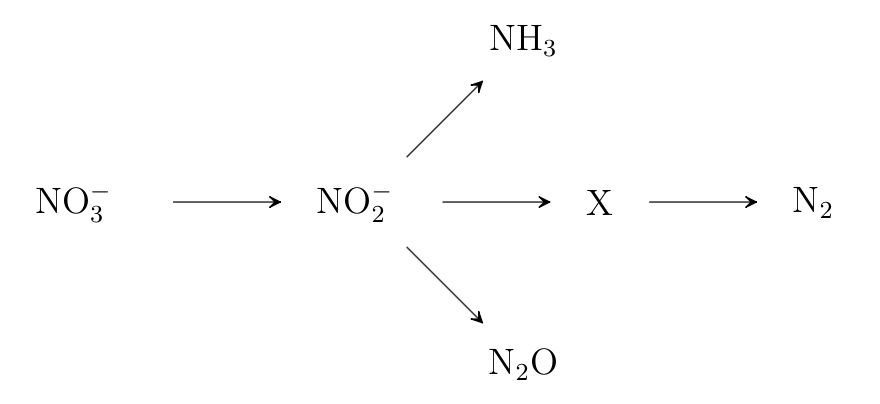}
\end{center}
\caption{Reaction kinetic model:
Simplified reaction scheme for the reduction of nitr4ate.}
\end{figure}

We consider the problem of estimating kinetic parameters of multi-step chemical
reactions that involve various intermediate or final products.
In particular, we study the dynamical system that describes the catalytic
conversion of nitrate ($\mbox{NO}_3^-$) to nitrogen ($\mbox{N}_2$) and other
by-products by electrochemical means.
The mechanism that is followed is complex and not well understood.
In this work, we use the simplified model proposed by~\cite{katsounaros},
which includes the production of nitrogen ($\mbox{N}_2$), ammonia
($\mbox{NH}_3$), and nitrous oxide ($\mbox{N}_2\mbox{O}$) as final products, as
 well as that of nitrite ($\mbox{NO}_2^-$) and an unknown species $\mbox{X}$ as reaction intermediates (see \fref{fig:reduction}).
% \begin{scheme}[!htb]
% \label{sc:reduction}
% \begin{center}
% \schemestart
%  $\rm NO_3^-$ \phantom{I}
%  \arrow(A--B){->}
%  $\rm NO_2^-$\phantom{I}
%  \arrow{->}
%  X
% \arrow(C--){->}
%  $\rm N_2$
%  \arrow(@B--C){->}[-45]
%  $\rm N_2O$
%  \arrow(@B--C){->}[45]
%  $\rm NH_3$
% \schemestop
% \end{center}
% \caption{Reaction kinetic model:
% Simplified reaction scheme for the reduction of nitrate.}
% \end{scheme}

The dynamical system associated with the reaction depicted in
\fref{fig:reduction} is:
\begin{equation}
\label{eqn:kinetic_model}
\begin{array}{cc}
\frac{d [\mbox{NO}_3^-]}{dt} &= -k_1[\mbox{NO}_3^-], \\
\frac{d[\mbox{NO}_2^-]}{dt} &= k_1[\mbox{NO}_3^-] - (k_2 + k_4 +
k_5)[\mbox{NO}_2^-], \\
\frac{d [\mbox{X}]}{dt} &= k_2 [\mbox{NO}_2^-] - k_3 [X],\\
\frac{d [\mbox{N}_2]}{dt} &= k_3 [\mbox{X}], \\
\frac{d [\mbox{NH}_3]}{dt} &= k_4 [\mbox{NO}_2^-],\\
\frac{d [\mbox{N}_2O]}{dt} &= k_5 [\mbox{NO}_2^-],
\end{array}
\end{equation}
where $[\cdot]$ denotes the concentration of a quantity, and
$k_i > 0$, $i=1,...5$ are the \emph{kinetic rate constants}.
Our goal is to estimate the kinetic rate constants based on the experimental
measurements obtained by~\cite{katsounaros}.
For completeness, these measurements are reproduced in \tref{table:catalysis_data}.
The initial conditions for \eqref{kinetic_model} are given by the $t=0$ row
of the \tref{table:catalysis_data}.

\begin{table}[!bth]
\centering
\begin{tabular}{c|cccccc}
%&i & 1 & 2 & 3 & 4 & 5 & 6 \\
\hline\hline
$t$ (min) & $[\mbox{NO}_3^-]$ & $[\mbox{NO}_2^-]$ & $[\mbox{X}]$ & $[\mbox{N}_2]$ &
$[\mbox{NH}_3]$ & $[\mbox{N}_2\mbox{O}]$ \\
\hline
0 & 500.00 & 0.00 & - & 0.00 & 0.00 & 0.00 \\
30 & 250.95 & 107.32 & - & 18.51 & 3.33 & 4.98 \\
60 & 123.66 & 132.33 & - & 74.85 & 7.34 & 20.14 \\
90 & 84.47 & 98.81 & - & 166.19 & 13.14 & 42.10 \\
120 & 30.24 & 38.74 & - & 249.78 & 19.54 & 55.98 \\
150 & 27.94 & 10.42 & - &292.32 & 24.07 & 60.65 \\
180 & 13.54 & 6.11 & - & 309.50 & 27.26 & 62.54 \\
\hline
\end{tabular}
\label{table:catalysis_data}
\caption{Reaction kinetic model:
The table contains the experimental data used in the calibration process.
The rows correspond to the time of each measurement and the columns to the
concentrations measured in $\mbox{mmol}\cdot\mbox{L}^{-1}$.
The ``-" symbols corresponds to lack of observations.
See~\cite{katsounaros} for more details on the experiment.
The $t=0$ row provides the initial condition to \eqref{kinetic_model}.
The observed data vector $\by \in\R^{30}$ is built by concatenating the scaled
version of the rows $t=30$ to $t=180$ while skipping the row corresponding to
$\mbox{X}$.
}
\end{table}

In order to avoid numerical instabilities, we work with a dimensionless
version of \eqref{kinetic_model}.
In particular, we define the scaled time:
\begin{equation}
\label{eqn:scaled_time}
\tau = \frac{t}{180\;\mbox{min}},
\end{equation}
the scaled concentrations:
\begin{equation}
\label{eqn:scaled_concentration}
u_i = \frac{[\mbox{Y}]}{500\;\mbox{mmol}\cdot\mbox{L}^{-1}},
\end{equation}
for $i=1,2,3,4,5, 6, 7$, and
$\mbox{Y} = \mbox{NO}_3^-, \mbox{NO}_2^-, \mbox{X}, \mbox{N}_2, \mbox{NH}_3, \mbox{N}_2\mbox{O}$, respectively,
and the scaled kinetic rate constants:
\begin{equation}
\label{eqn:scaled_rate}
\kappa_i = k_i \cdot 180\;\mbox{min},
\end{equation}
for $i=1,\dots,5$.
The dimensionless dynamical system associated with \eqref{kinetic_model} is:
\begin{equation}
\label{eqn:scaled_kinetic_model}
\begin{array}{cc}
\dot{u}_1 &= -\kappa_1 u_1, \\
\dot{u}_2 &= \kappa_1 u_1 - (\kappa_2 + \kappa_4 + \kappa_5)u_2, \\
\dot{u}_3 &= \kappa_2 u_2 - \kappa_3 u_3,\\
\dot{u}_4 &= \kappa_3 u_3, \\
\dot{u}_5 &= \kappa_4 u_2,\\
\dot{u}_6 &= \kappa_5 u_2,
\end{array}
\end{equation}
where $\dot{u}$ denotes the differentiation of $u$ with respect to the
scaled time $\tau$.
The initial conditions for \eqref{scaled_kinetic_model} are given by the scaled
version of the $t=0$ row of \tref{table:catalysis_data}.
We arrange the scaled version of the experimental data of
\tref{table:catalysis_data} in a vector form, $\by\in\R^{d_y}$ with $d_y=30$,
by concatenating rows $t=30,\dots,180$ of \tref{table:catalysis_data} while skipping
the third column.

\begin{table}[!htb]
\caption{Reaction kinetic model:
The logarithm of the scaled kinetic rate constants, $\xi_i$ (see
\eqref{log_kappa_i}, and the logarithm of the likelihood noise,
$\theta$ (see \eqref{iso_gauss_like}) as estimated by the variational approach
with $L=1$ and MCMC (MALA).
The estimates correspond to the mean and the uncertainties to two times the
standard deviation of each method.
}
\label{table:rate_constants}
\begin{center}
\begin{tabular}{c r r}
\hline \hline
Variable & VAR ($L=1$) & MCMC (MALA) \\
\hline
$\xi_1$ & $1.359 \pm 0.055$ & $1.356 \pm 0.072$ \\
$\xi_2$ & $1.657 \pm 0.086$ & $1.664 \pm 0.142$ \\
$\xi_3$ & $1.347 \pm 0.118$ & $1.349 \pm 0.215$ \\
$\xi_4$ & $-0.162 \pm 0.167$ & $-0.159 \pm 0.230$ \\
$\xi_5$ & $-1.009 \pm 0.368$ & $-1.071 \pm 0.513$ \\
$\theta$ & $-3.840 \pm 0.204$ & $-3.757 \pm 0.251$ \\
\hline
\end{tabular}
\end{center}
\end{table}

%\begin{table}[!htb]
%\caption{Reaction kinetic model:
%{\bf This shows the 95\% credible intervals. Should we use it instead of
%\tref{table:rate_constants} or not?}
%The logarithm of the scaled kinetic rate constants, $\xi_i$ (see
%\eqref{log_kappa_i}, and the logarithm of the likelihood noise,
%$\theta$ (see \eqref{iso_gauss_like}) as estimated by the variational approach
%with $L=1$ and MCMC (MALA).
%The estimates correspond to the mean and the uncertainties to two times the
%standard deviation of each method.
%}
%\label{table:rate_constants3}
%\begin{center}
%\begin{tabular}{c r r r r}
%\hline \hline
%Variable & VAR Median & VAR $95\%$ Int. & MCMC Median & MCMC $95\%$ Int. \\
%\hline
%$\xi_1$ & $1.359$ & $(1.313, 1.405)$ & $1.356$ & $(1.298, 1.419)$ \\
%$\xi_2$ & $1.657$ & $(1.589, 1.723)$ & $1.664$ & $(1.554, 1.785)$ \\
%$\xi_3$ & $1.347$ & $(1.255, 1.441)$ & $1.349$ & $(1.182, 1.536)$ \\
%$\xi_4$ & $-0.162$ & $(-0.308, -0.016)$ & $-0.159$ & $(-0.357, 0.017)$ \\
%$\xi_5$ & $-1.009$ & $(-1.329, -0.692)$ & $-1.071$ & $(-1.506, -0.713)$ \\
%$\theta$ & $-3.840$ & $(-4.006, -3.675)$ & $-3.757$ & $(-3.960, -3.552)$ \\
%\hline
%\end{tabular}
%\end{center}
%\end{table}

Since the kinetic rate constants, $\kappa_i$, of the scaled dynamical system
of \eqref{scaled_kinetic_model} are non-negative, it is problematic to attempt
to approximate the posteriors associated with them with mixtures of Gaussians.
For this reason, we work with the logarithms of the $\kappa_i$'s,
\begin{equation}
\label{eqn:log_kappa_i}
\xi_i = \log \kappa_i,
\end{equation}
for $i=1,\dots,5$.
We collectively denote those variables by $\bxi = (\xi_1,\dots,\xi_5)$.
The prior probability density we assign to $\bxi$ is:
\begin{equation}
\label{eqn:catalysis_prior_xi}
p(\bxi) = \prod_{i=1}^{5}p(\xi_i),
\end{equation}
with
\begin{equation}
\label{eqn:catalysis_prior_xi_i}
p(\xi_i) = \calN(\xi_i | 0, 1),
\end{equation}
for $i=1,\dots,5$.

\begin{table}[hbt]
\centering
\begin{tabular}{clll}
\hline\hline
Rate constant & Katsounaros (2012) & VAR Median & VAR 95\% Interval\\
\hline
$k_1$ & $0.0216 \pm 0.0014$ & 0.0216 & (0.0205, 0.0229) \\
$k_2$ & $0.0292 \pm 0.0036$ & 0.0291 & (0.0269, 0.0316) \\
$k_3$ & $0.0219 \pm 0.0044$ & 0.0214 & (0.0191, 0.0239) \\
$k_4$ & $0.0021 \pm 0.0008$ & 0.0020 & (0.0014, 0.0030) \\
$k_5$ & $0.0048 \pm 0.0008$ & 0.0047 & (0.0040, 0.0056) \\
\hline
$\sigma$ & Not available & 0.0215 & (0.0176, 0.0262) \\
\hline
\end{tabular}
\label{table:vs_katsounaros}
\caption{
Reaction kinetic model:
The first five rows compare the median and
$95\%$ credible intervals of the kinetic rate constants estimated via the
variational approach to those found in~\cite{katsounaros}.
The units of the rates are in $\mbox{min}^{-1}$.
The last line shows the median and $95\%$ credible interval of the scaled
measurement noise $\sigma = e^{\theta}$.
This quantity is unit-less.}
\end{table}

The forward model $\bff:\R^{5}\rightarrow\R^{30}$ associated with the
experimental observations $\by$ is:
\begin{equation}
\bff(\bxi) =
\left(
\begin{array}{c}
u_1(t_2, \bxi),  u_2(t_2, \bxi),u_4(t_2, \bxi) u_5(t_2,\bxi),u_6(t_2,\bxi),\\
\dots\\
u_1(t_7, \bxi),  u_2(t_7, \bxi),u_4(t_7, \bxi) u_5(t_7,\bxi),u_6(t_7,\bxi),
\end{array}
\right),
\end{equation}
where $u_i(t,\bxi)$ is the solution of \eqref{scaled_kinetic_model} with the
initial conditions specified by the scaled version of the $t=0$ row of
\tref{table:catalysis_data}, and scaled kinetic rate constants, $\kappa_i$,
given by inverting \eqref{log_kappa_i}, i.e. $\kappa_i = e^{\xi_i}$.
The derivatives of $\bff(\bxi)$ can be solving a series of dynamical systems forced by the solution of \eqref{scaled_kinetic_model}.
This is discussed in~\ref{ap:dyn_adjoint}.

\begin{figure}[!htb]
\label{fig:catalysis_input}
\centering
\includegraphics[width=65mm]{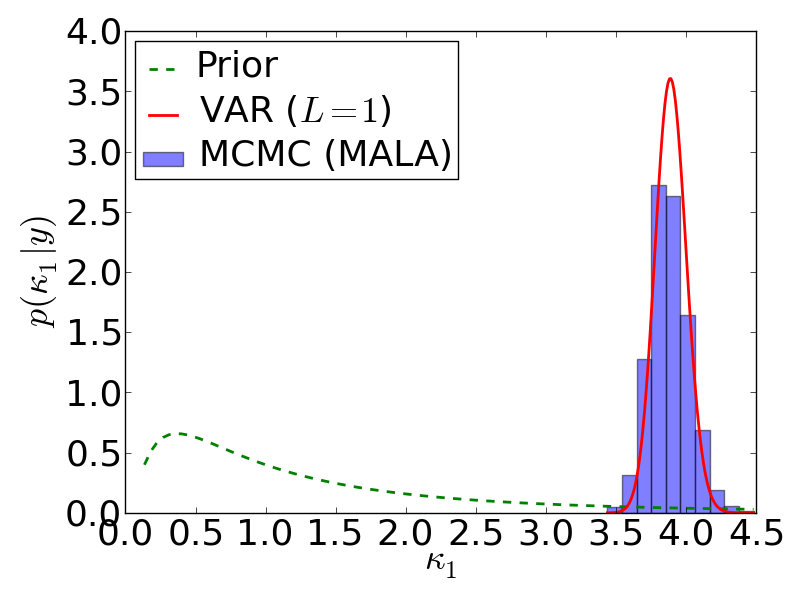}
\includegraphics[width=65mm]{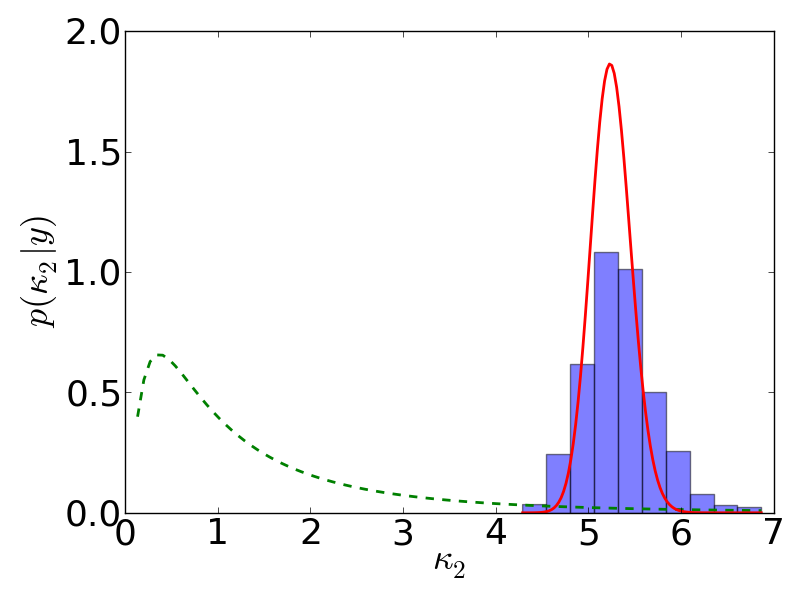}
\includegraphics[width=65mm]{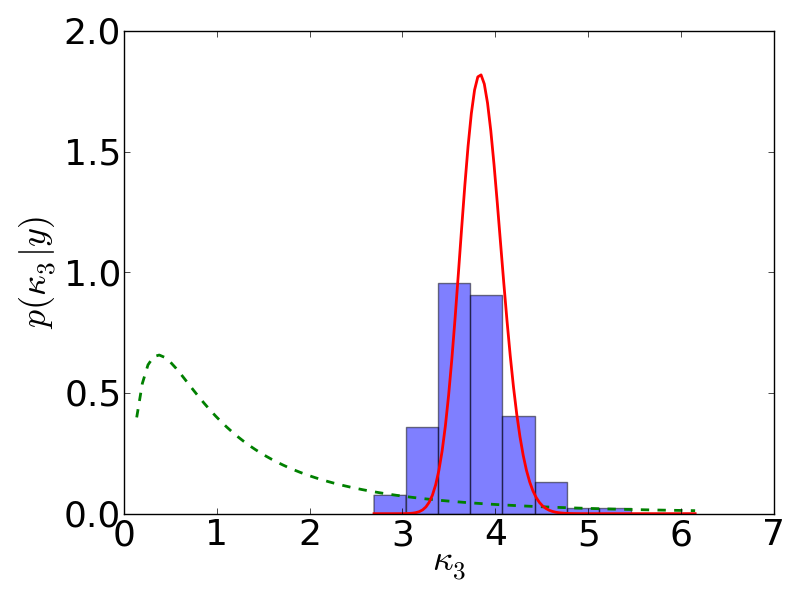}
\includegraphics[width=65mm]{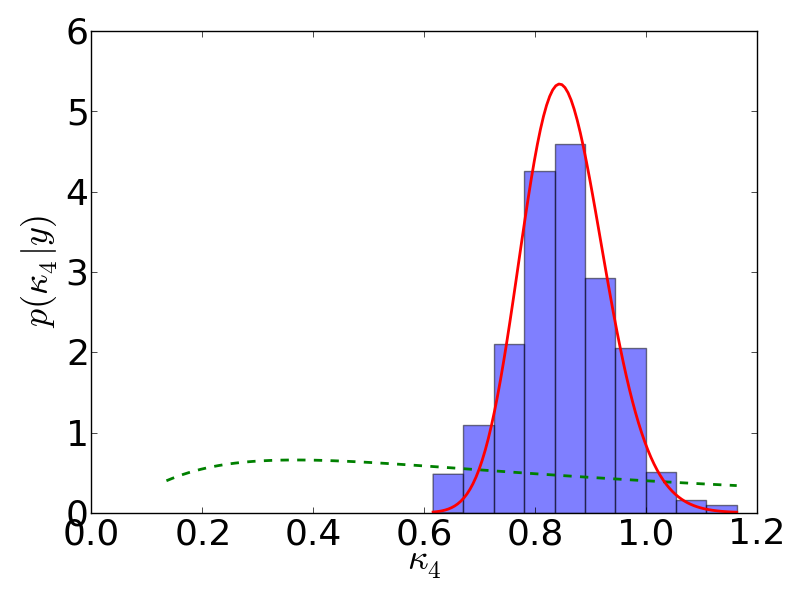}
\includegraphics[width=65mm]{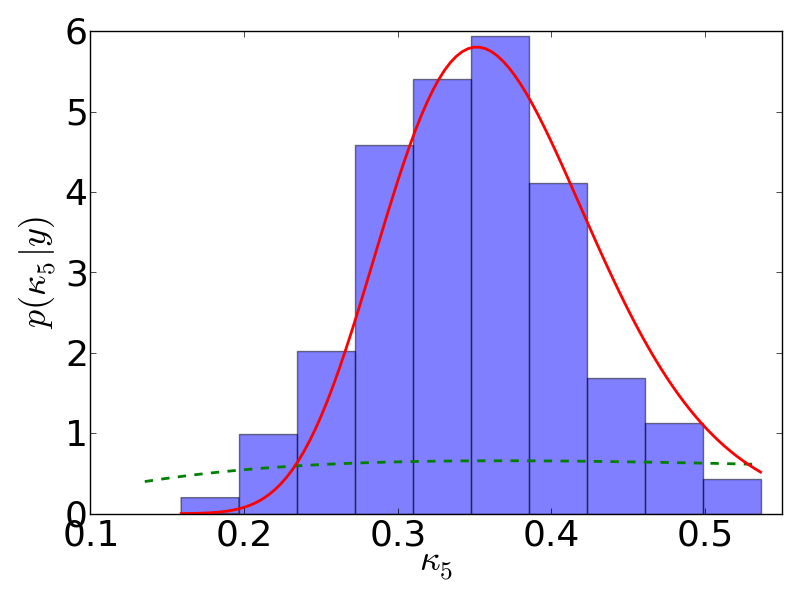}
\includegraphics[width=65mm]{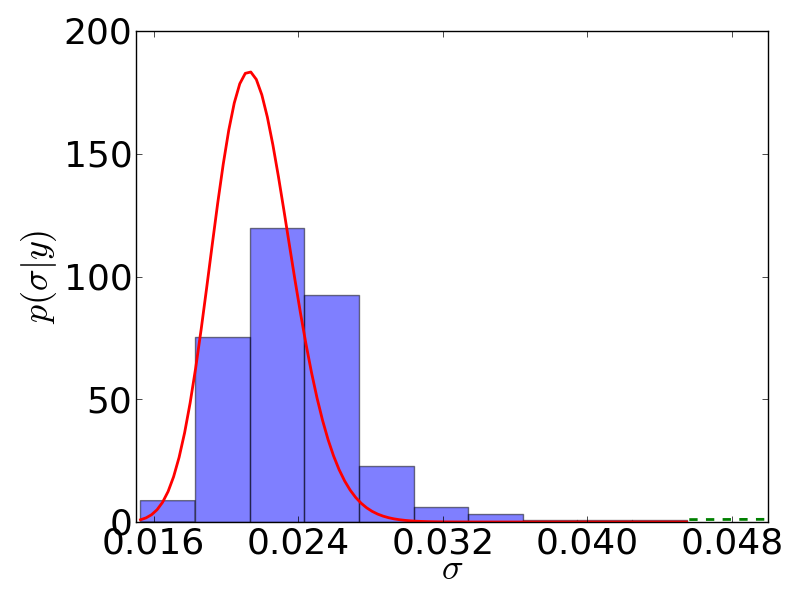}
\caption{Reaction kinetic model:
Comparison of the variational posterior (VAR ($L=1$)) of the scaled kinetic rate
constants $\kappa_i$ as well as of the likelihood noise $\sigma$ to the MCMC
(MALA) histograms of the same quantity.
The prior probability density of each quantity is shown as a dashed green line.}
\end{figure}

We use the isotropic Gaussian likelihood defined in
\eqref{iso_gauss_like}.
It is further discussed in \ref{ap:iso_gauss_like}.
The prior we assign to the parameter $\theta$ of the likelihood is:
\begin{equation}
\label{eqn:prior_theta}
p(\theta) = \calN(\theta | -1, 1).
\end{equation}
Since the noise represented by $\theta$ is $\sigma = e^{\theta}$, this prior
choice corresponds to a belief that the measurement noise is around $30$\%.

\begin{figure}[!htb]
\label{fig:catalysis_uq}
\begin{center}
\includegraphics[width=65mm]{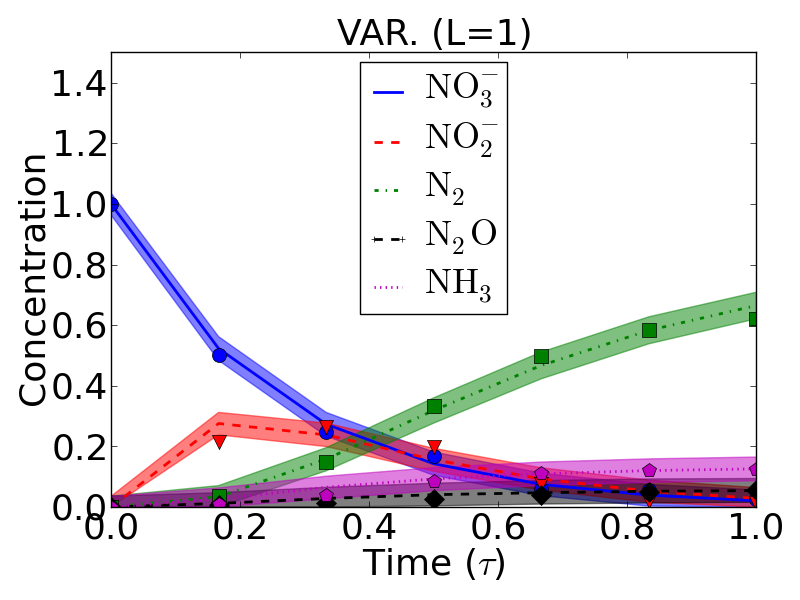}
\includegraphics[width=65mm]{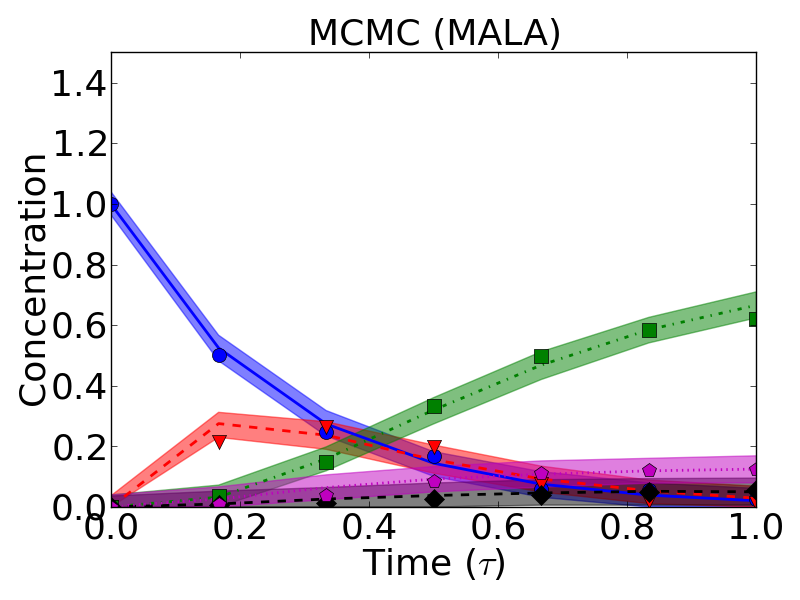}
\end{center}
\caption{Reaction kinetic model:
The `$\bullet$', `$\triangledown$', `$\square$', `$\diamond$', and
`$\pentagon$' signs indicate the scaled experimental measurements for
$\mbox{NO}_3^-$, $\mbox{NO}_2^-$, $\mbox{N}_2$, $\mbox{N}_2\mbox{O}$,
and $\mbox{NH}_3$, respectively.
The lines and the shaded areas around them correspond to the median
and $95\%$ credible intervals of the scaled concentration, $u_i$, as a function of the scaled time $\tau$.
The left plot, shows the results obtained by approximating the posterior of
the parameters via the variational approach.
The right plot, shows the results obtained via MCMC (MALA).}
\end{figure}

We solve the problem using the variational approach as outlined in \aref{vi}.
To approximate the posterior, we only use one Gaussian, $L=1$ in \eqref{mixture}.
We impose no bounds on $\bmu_1 = \bmu\in\R^d$.
However, we require that the diagonal elements $\sigma_i^2$ of the covariance
matrix $\bSigma_1 = \bSigma = \operatorname{diag}\left(\sigma_1,\dots,\sigma_d \right)$ are bounded below by $10^{-6}$ and above by $10^2$.

\Tref{table:rate_constants} compares the variational estimates of the scaled
kinetic rate constants, $\xi$ (see \eqref{log_kappa_i}), and the logarithm of
the likelihood noise, $\theta$ (see \eqref{iso_gauss_like}), to the MCMC (MALA)
estimates.
We see that the mean of the two estimates are in close agreement,
albeit the variational approach slightly underestimates the uncertainty
of its prediction.
However, notice that if we order the parameters in terms of increasing
uncertainty, both methods yield the same ordering.
Therefore, even though the numerical estimates of the uncertainty differ, the
relative estimates of the uncertainty are qualitatively the same.
It is worth mentioning at this point, that the variational approach uses only
$37$ evaluations of the forward model.
This is to be contrasted with the thousands of evaluations required so that
the MCMC (MALA) estimates converge.

The variational approach with $L=1$ approximates the posterior of $\bxi$ and
$\theta$ with one multivariate Gaussian distribution with a diagonal covariance.
Therefore, the distribution of each one of the components is a Gaussian.
Using Eqs.~(\ref{eqn:log_kappa_i}) and~(\ref{eqn:scaled_rate}), it is easy
to show that the kinetic rate constant $k_i$ follows a log-normal
distribution with log-scale parameter $\mu_i - \log(180)$ and shape parameter
$\sigma_i$.
Similarly, we can show that the noise $\sigma = e^{\theta}$ follows
a log-normal distribution with local-scale parameter $\mu_d$ and shape parameter
$\sigma_d$.
Using the percentiles of these lognormal distributions, we compute the median
and the $95\%$ credible intervals of the kinetic rate constants $k_i$ and
the noise $\sigma$.
The results are shown in the third and fourth columns of \tref{table:vs_katsounaros}.
They are in good agreement with the results found in~\cite{katsounaros} using
a MCMC strategy (reproduced in the second column of the same
\tref{table:vs_katsounaros}).
An element of our analysis not found in~\cite{katsounaros} is the estimation of
the measurement noise.
Since $\sigma$ measures the noise of the scaled version of the data $\by$,
we see (last line of \tref{table:vs_katsounaros}) that the measurement noise is
estimated to be around 2.15\%.

In \fref{fig:catalysis_input}, we compare the variational posterior (VAR ($L=1$))
of the scaled kinetic rate constants, $\kappa_i$, as well as of
the noise of the likelihood, $\sigma$, to histograms
of the same quantities obtained via MCMC (MALA).
Once again, we confirm the excellent agreement between the two methods.
In the same figure, we also plot the prior probability density we assigned to
each parameter.
The prior probability of $\sigma$ is practically invisible, because it picks
at about $\sigma = 0.30$.
Given the big disparity between prior and posterior distributions,
we see that the result is relatively insensitive to the priors we assign.
If, in addition, we take into account that the measurement noise is estimated
to be quite small, we conclude that \eqref{kinetic_model} does a very good job of explaining
the experimental observations.

\Fref{fig:catalysis_uq} shows the uncertainty in the scaled concentrations,
$u_i$, as a function of scaled time, $\tau$, obtained by approximating the
posterior of the parameters and compares them with the variational approach (left), to the
MCMC (MALA) estimate.
Again, we notice that the two plots are in very good agreement, albeit the
variational approach seems to slightly underestimate the uncertainty.

\subsection{Contaminant source identification}
\label{sec:contamination}
We now apply our methodology to a synthetic example of contaminant source
identification.
We are assuming that we have experimental measurements of contaminant
concentrations at specific locations, and we are interested in estimating
the location of the contamination source.

\begin{figure}[htb]
\label{fig:diffusion_1}
\begin{center}
\includegraphics[width=65mm]{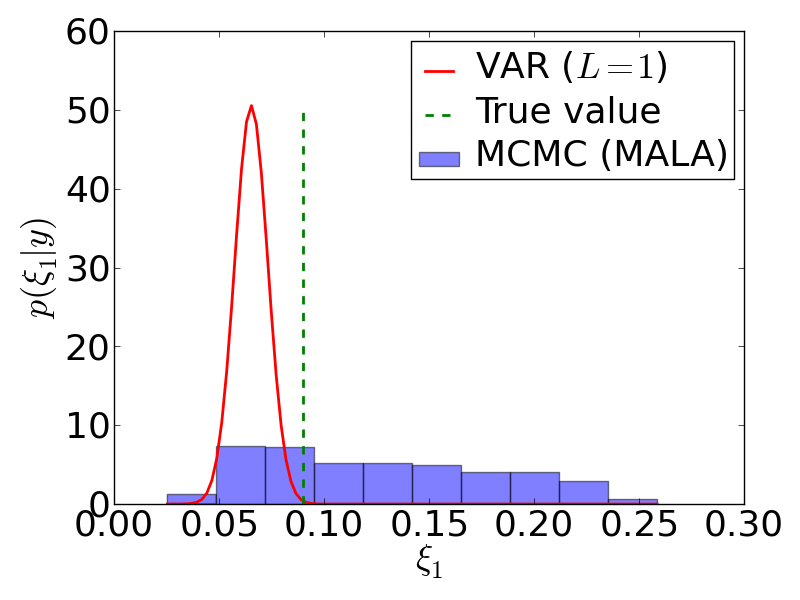}
\includegraphics[width=65mm]{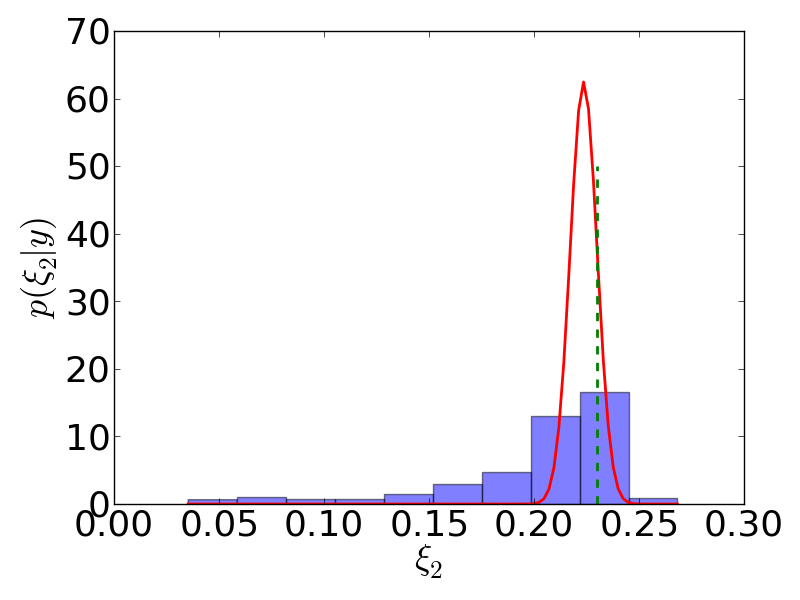}
\includegraphics[width=65mm]{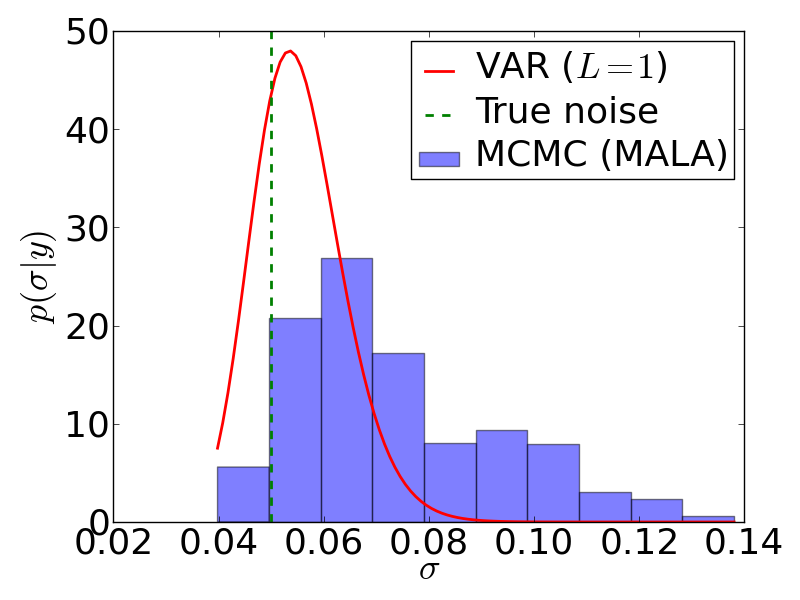}
\end{center}
\caption{Contaminant source identification, first case:
Comparison of the variational posterior (VAR ($L=1$)) of the source location
$\bxi = (\xi_1,\xi_2)$ as well as of the likelihood noise $\sigma$ to the MCMC
(MALA) histograms of the same quantities.
The true value of each quantity is marked by a vertical, green, dashed line.
}
\end{figure}

The concentration of the contaminant obeys the two-dimensional transport model
described by a diffusion equation
\begin{eqnarray}
\label{eqn:transport}
\frac{\partial u}{\partial t} = \nabla^2 u + g(t,\bx, \bxi),\;\bx \in B,
\end{eqnarray}
where $B = [0,1]^2$ is the spatial domain and $g(t,\bx, \bxi)$ is the source
term.
The source term is assumed to have a Gaussian:
\begin{eqnarray}
g(t, \bss; \bx_s) = g_0 e^{-\frac{|\bx -
   \bxi|^2}{2\rho^2}} 1_{[0, T_s]}(t),
\end{eqnarray}
where $g_0 = \frac{1}{\pi\rho}$ is the strength of the contamination, $\rho=0.05$ is its spreadwidth,  $T_s = 0.3$ is the shutoff
time parameter, and $\bxi$ is the source center.
Therefore, $\bxi$ is the only parameter that needs to be identified
experimentally.
We impose homogeneous Neumann boundary conditions
\begin{eqnarray}
\nabla \bu \cdot \mathbf{n} = 0, \;\bx \in \partial B
\end{eqnarray}
and zero initial condition
\begin{eqnarray}
\bu(0, \bx, \bxi) = 0.
\end{eqnarray}

\begin{figure}[htb]
\label{fig:two_comp}
\begin{center}
\includegraphics[width=65mm]{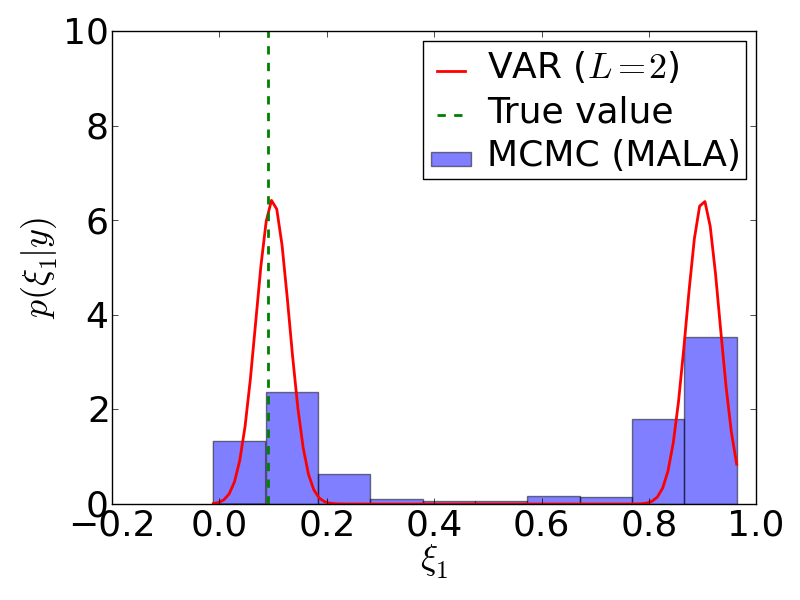}
\includegraphics[width=65mm]{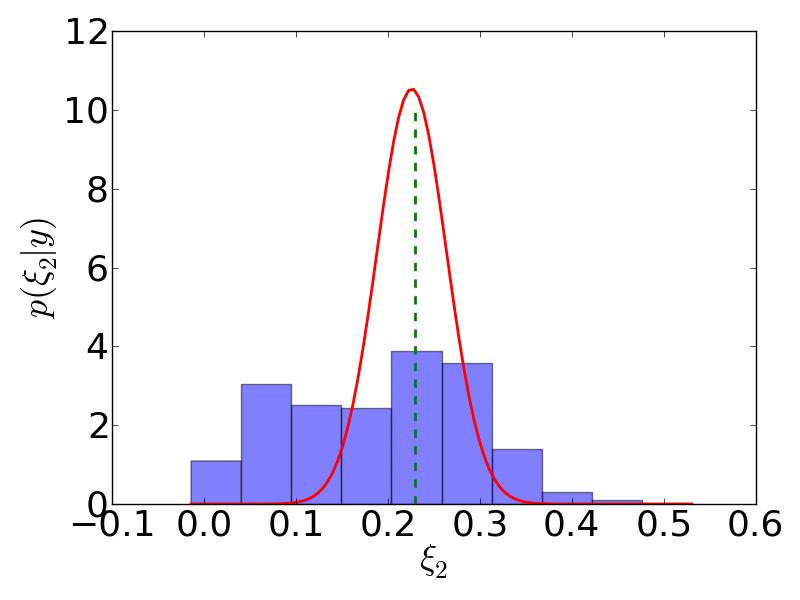}
\includegraphics[width=65mm]{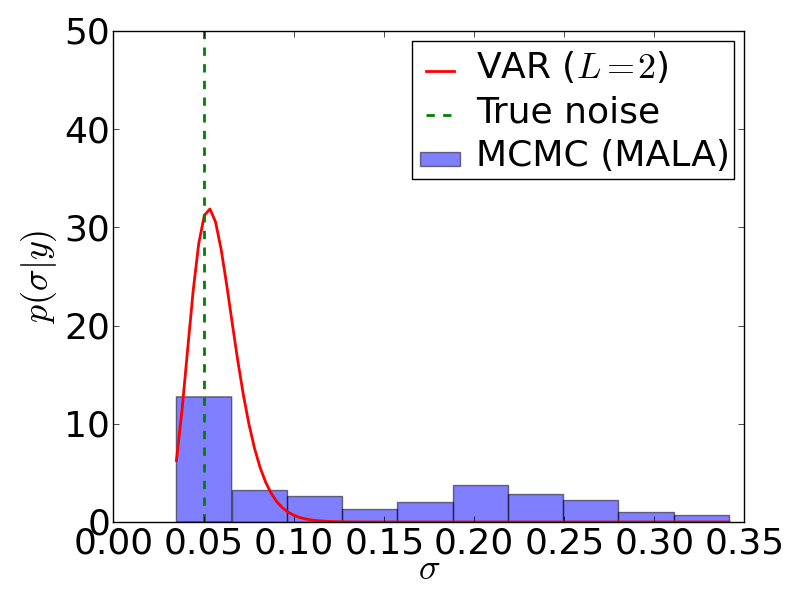}
\end{center}
\caption{Contaminant source identification, second case:
Comparison of the variational posterior (VAR ($L=1$)) of the source location
$\bxi = (\xi_1,\xi_2)$ as well as of the likelihood noise $\sigma$ to the MCMC
(MALA) histograms of the same quantities.
The true value of each quantity is marked by a vertical, green, dashed line.
}
\end{figure}

We consider two different scenarios.
In the first one, measurements of $u(t, \bx, \bxi)$ take place on the four
corners of $B$.
On the second one, measurements of $u(t, \bx, \bxi)$ take place on the middle
points of the upper and lower boundaries of $B$.
The former results in a unimodal posterior for $\bxi$ and can be approximated
with just one Gaussian.
The latter results in a bimodal posterior for $\bxi$ and requires a
mixture of two Gaussians.
\eqref{transport} is solved via a finite volume scheme implemented using the
Python package Fipy\cite{fipy}.
The required gradients of the solution $u(t,\bx,\bxi)$ are obtained by solving
a series of PDE's similar to \eqref{transport} but with different source
terms (see \ref{ap:adjoint_diffusion}).
In both scenarios, we generate synthetic observations, $\by$, by solving
\eqref{transport} on a $110\times 110$ grid,
source $\bxi^* = (0.09, 0.23)$, and adding Gaussian noise with standard
deviation $\sigma^* = 0.05$.
For the forward evaluations needed during
the calibration process we use a $25 \times 25$ grid and we denote the
corresponding solution by $\tilde{\bu}(t, \bss; \bxi)$.
The prior we use for $\bxi$ is uniform,
\begin{equation}
\label{eqn:priorsource}
p(\bxi) \propto 1_B(\bxi),
\end{equation}
the likelihood is given by \eqref{iso_gauss_like}, and the log-noise parameter
$\theta$ of the likelihood has the same prior as the previous example
(see \eqref{prior_theta}).

\paragraph{First case: Observations at the four corners}
The synthetic data, $\by\in\R^{16}$ ($4$ sensors $\times$ $4$ measurements)
are generated by sampling the $110\times 110$
solution, $u(t, \bx) := u(t, \bx, \bxi^*)$ at the four corners of $B$, and by adding
Gaussian noise with $\sigma^* = 0.05$:
\begin{equation}
\begin{array}{ccc}
\by &=& \left( \begin{array}{c} u(t_1, (0,0)), u(t_1,(1,0)), u(t_1,
    (0,1)), u(t_1, (1,1)), \\ \vdots \\  u(t_4, (0,0)), u(t_4, (1,0)),
    u(t_4, (0,1)), u(t_4, (1,1)) \end{array} \right)\\
    && +\;\mbox{noise}.
\end{array}
\end{equation}
The corresponding forward model generated by the $25\times 25$ solution,
$\tilde{u}(t,\bx,\bxi)$, is given by:
\begin{equation}
\bff(\bxi) =  \left( \begin{array}{c} \tilde{u}(t_1, (0,0), \bxi),
    \tilde{u}(t_1, (1,0), \bxi), \tilde{u}(t_1, (0,1), \bxi),
    \tilde{u}(t_1, (1,1), \bxi), \\ \vdots \\  \tilde{u}(t_4, (0,0),
    \bxi), \tilde{\bu}(t_4, (1,0), \bxi), \tilde{u}(t_4, (0,1); \bxi),
    \tilde{u}(t_4, (1,1), \bxi) \end{array} \right).
\end{equation}

\Fref{fig:diffusion_1} compares the posteriors obtained via the
variational approach with $L=1$ Gaussian in \eqref{mixture} to those obtained
via MCMC (MALA).
The true value of each parameter is indicated by a vertical, green, dashed line.
It is worth noting at this point,
that the variational approach required $48$
forward model evaluations as opposed to thousands required by MCMC (MALA).
In real time, the variational approach took about 15 minutes on a single
computational node, while the MCMC (MALA) required 3 days on the same node.
Notice that the posterior cannot identify the true source exactly.
This is due to the $5\%$ noise that we have added in the synthetic data.
The result of such noise is always to broaden the posterior.
We see that the variational approach does a good job in identifying an
approximate location for the source $\bxi$ as well as estimating the noise
level $\sigma$.
However, we notice once again that it underestimates the true uncertainty.

\paragraph{Second case: Observations at the middle point of the upper and lower boundaries}
The synthetic data, $\by\in\R^{8}$ ($2$ sensors $\times$ $4$ measurements)
are generated by sampling the $110\times 110$
solution, $u(t, \bx) := u(t, \bx, \bxi^*)$ at the upper and lower boundaries
of $B$, and by adding
Gaussian noise with $\sigma^* = 0.05$:
\begin{equation}
\begin{array}{ccc}
\by &=& \left( \begin{array}{c} u(t_1, (0.5,0)), u(t_1,(0.5,1)), \\ \vdots \\  u(t_4, (0.5,0)), u(t_4, (0.5,1)) \end{array} \right)\\
    && +\;\mbox{noise}.
\end{array}
\end{equation}
The corresponding forward model generated by the $25\times 25$ solution,
$\tilde{u}(t,\bx,\bxi)$, is given by:
\begin{equation}
\bff(\bxi) =  \left( \begin{array}{c} \tilde{u}(t_1, (0.5,0), \bxi),
    \tilde{u}(t_1, (0.5,1), \bxi), \\ \vdots \\  \tilde{u}(t_4, (0.5,0),
    \bxi), \tilde{\bu}(t_4, (0.5,1), \bxi) \end{array} \right).
\end{equation}

\Fref{fig:two_comp} compares the posteriors obtained via the
variational approach with $L=2$ Gaussians in \eqref{mixture} to those obtained
via MCMC (MALA).
The true value of each parameter is indicated by a vertical, green, dashed line.
It is worth noting at this point,
that the variational approach required only $62$
forward model evaluations.
Using symmetry arguments, it is easy to show that
data generated by solving \eqref{transport}
with a source located at $(\xi_1, \xi_2)$
look identical to the data that can be generated from a source located at
$(1-\xi_1, \xi_2)$.
As a result, the posterior distribution is bimodal.
Therefore, we expect common MCMC methodologies to have a hard time dealing
with this problem.
The reason is that once the MCMC chain visits one of the modes, it is very
unlikely that it will ever leave it to visit the other mode.
In reality, it is impossible to visit the other mode unless a direct jump is
 proposed.
The reason our MCMC (MALA) scheme works is because we have handpicked a proposal
step that does allow for occasional jumps from one mode to the other.
On the other hand, we see that the variational approach with $L=2$ Gaussians in
\eqref{mixture} can easily deal with bimodal (or multimodal) posteriors.
However, there are a few details that need to be mentioned here.
Firstly, one needs to use an $L$ greater than or equal to the true number of
modes of the posterior.
Since, the latter is unknown, a little bit of experimentation would be required
in a real problem.
Secondly, even if the true $L$ is used, \aref{vi} might still find fewer modes
than the true number.
For example, in our numerical experiments we have noticed that if $\bmu_1$ and
$\bmu_2$ of \eqref{mixture} with $L=2$ are initially very close together, they are both attracted by the same mode.
We believe that this is an artifact introduced by the Taylor approximation to
the joint probability function $\calL[q]$ (see \eqref{exp_term_0} and
\eqref{exp_term_1}), and, in particular, of its local nature.
In our example, a few random initializations of the $\bmu_i$'s are enough to
guarantee the identification of both posterior modes. 

\section{Conclusions}
\label{sec:conclusions}

We presented a novel approach to inverse problems that provides an optimization
perspective to the Bayesian point of view.
In particular, we used information theoretic arguments to derive an optimization
problem that targets the estimation of the posterior within the the class
of mixtures of Gaussians.
The scheme proceeds by postulating that the ``best" approximate posterior is the
one that exhibits the minimum information loss (relative entropy) within the
class of candidate posteriors.
We showed how the minimization of the information loss is equivalent to the
maximization of a lower bound to the evidence (normalization constant of the
posterior).
Since the derived lower bound was a computationally intractable quantity, we
derived a crude approximation to it that requires the gradients of the forward
model with respect to the input variables that we want to infer. 

We demonstrated the efficacy of our method to
solve inverse problems with just a few forward model evaluations in two
numerical examples: 1) the estimation of the kinetic rate constants in a
catalysis system, and 2) the identification of the contamination source in a
simple diffusion problem.
The performance of the scheme was compared to that of a state of the art MCMC
technique (MALA) and was found to be satisfactory, albeit slightly
underestimating the uncertainty.
The scheme was able to solve both inverse problems with a fraction of the
computational cost.
In particular, our approach required around 50 forward model evaluations
as opposed to the tens of thousands that are required by MCMC. 

The variational approach seems to open up completely new ways of solving stochastic
inverse problems.
The scope of the approach is much wider than the particular techniques used in this paper.
Just as a indication, the following are some of the research directions that we plan to pursue in the near
future: 1) Derive alternative -more accurate- approximations to the lower bound of the evidence;
2) Experiment with dimensionality reduction ideas that would allow us to carry out the variational optimization in high-dimensional problems;
3) Derive stochastic algorithms for maximizing the lower bound without invoking any approximations.
We believe that the variational approach has the potential of making stochastic inverse problems solvable
with only a moderate increase in the computational cost as compared to classical approaches.

\section*{Acknowledgments}

I.K. acknowledges financial support through a Marie Curie International
Outgoing Fellowship within the 7th European Community Framework Programme
(Award IOF-327650).
N.Z, as `Royal Society Wolfson Research Merit Award'
holder acknowledges support from the Royal Society and
the Wolfson Foundation. N.Z. also acknowledges strategic
grant support from EPSRC to the University of Warwick
for establishing the Warwick Centre for Predictive
Modeling (grant EP/L027682/1). In addition, N.Z.
as Hans Fisher Senior Fellow acknowledges support of
the Technische Universit$\ddot{a}$t M$\ddot{u}$nchen - Institute for Advanced
Study, funded by the German Excellence Initiative
and the European Union Seventh Framework Programme
under grant agreement no. $291763$.

\appendix

\section{Computation of $\calF_a[q]$ and its gradient}
\label{ap:F_grad}

\aref{vi} requires the evaluation of the gradient of the
approximate ELBO of \eqref{elbo} with respect to all the parameters of the
Gaussian mixture $q(\bomega)$ of \eqref{mixture}.
That is, we must be able to evaluate
\begin{eqnarray}
\calF_a[q] &=& \calH_0[q] + \calL_a[q],\\
\label{eqn:F_grad}
\frac{\partial}{\partial \beta} \calF_a[q] &=& \frac{\partial}{\partial \beta}\calH_0[q]
+ \frac{\partial}{\partial\beta}\calL_a[q],
\end{eqnarray}
for $\beta = w_i, \mu_{ij} = (\bmu_i)_{j}$, $\Sigma_{ijk} = (\bSigma_i)_{jk}$
for $i=1,\dots,L$, and $j,k=1,\dots,d$ and $a = 0,2$.

\subsection{Computation of $\calH_0[q]$ and its gradient}
\label{ap:H_compute}
The computations relative to $\calH_0[q]$ are:
\begin{eqnarray}
\calH_0[q] &=& -\sum_{i=1}^L w_i \log(q_i),\\
\frac{\partial}{\partial w_i}\calH_0[q] &=&
    -\log(q_i) - \sum_{r=1}^L\frac{w_rN_{ri}}{q_r},\\
\frac{\partial}{\partial \mu_{ij}}\calH_0[q] &=&
    -w_i \sum_{r=1}^L w_r N_{ri}A_{rij}\left(\frac{1}{q_i} + \frac{1}{q_r}\right),\\
\frac{\partial}{\partial \Sigma_{ijk}}H_0[q] &=&
\frac{1}{2}w_i \sum_{r=1}^L w_r N_{ri} B_{rijk}\left(\frac{1}{q_i} + \frac{1}{q_r}\right),
\end{eqnarray}
where, in order to simplify the notation, we have used the following
intermediate quantities:
\begin{eqnarray}
N_{ri} &=& \calN\left(\bmu_r | \bmu_i, \bSigma_r + \bSigma_i\right),\\
q_i &=& \sum_{r=1}^L w_r N_{ri},\\
A_{rij} &=& \sum_{s=1}^d\left(\bSigma_r + \bSigma_i\right)^{-1}_{js}\left(\mu_{rs} - \mu_{is} \right),\\
B_{rijk} &=& \left(\bSigma_r + \bSigma_i \right)^{-1}_{jk} + A_{rij}A_{rik}.
\end{eqnarray}

\subsection{Computation of $\calL_a[q]$ and its gradients}
\label{ap:L_compute}
The computations relative to the $\calL_a[q]$ part are:
\begin{eqnarray}
\calL_0[q] &=& \sum_{i=1}^L w_i C_i,\\
\frac{\partial}{\partial w_i}\calL_0[q] &=& C_i,\\
\frac{\partial}{\partial \mu_{ij}}\calL_0[q] &=& D_{ij},\\
\frac{\partial}{\partial \Sigma_{ijk}}\calL_0[q] &=& 0,\\
\label{eqn:L_2_c_1}
\calL_2[q] &=& \calL_0[q] + \frac{1}{2}\sum_{i=1}^L w_i \sum_{j,k=1}^d \Sigma_{ijk}E_{ijk},\\
\label{eqn:L_2_c_2}
\frac{\partial}{\partial w_i}\calL_2[q] &=& \frac{\partial}{\partial w_i}\calL_0[q]
+ \frac{1}{2}\sum_{j,k=1}^d \Sigma_{ijk}E_{ijk},\\
\label{eqn:L_2_c_3}
\frac{\partial}{\partial \Sigma_{ijk}}\calL_2[q] &=&
\frac{\partial}{\partial \Sigma_{ijk}}\calL_0[q]
+ \frac{1}{2}w_i E_{ijk},
\end{eqnarray}
where, in order to simplify the notation, we have used the following
intermediate quantities:
\begin{eqnarray}
\label{eqn:J}
J(\bomega) &=& \log p(\by, \bomega) = \log p(\by | \bomega) + \log p(\bomega),\\
C_i &=& J(\bmu_i),\\
\label{eqn:J_1}
D_{ij} &=& \frac{\partial }{\partial \omega_{j}}J(\bmu_i),\\
\label{eqn:J_2}
E_{ijk} &=& \frac{\partial^2}{\partial \omega_{j}\omega_{k}}J(\bmu_i).
\end{eqnarray}
We do not provide the derivatives of $\calL_2[q]$ with respect to
$\mu_{ijk}$ because they are not needed in \aref{vi}.
The joint probability function $J(\bomega)$ of \eqref{J} depends on the
details of the likelihood, the prior, and the underlying forward model.
We discuss the computation of Eqs.~(\ref{eqn:J}),~(\ref{eqn:J_1}), and~(\ref{eqn:J_2}) in \ref{ap:J_compute}.

\subsection{Computing the derivatives of $J(\bomega)$}
\label{ap:J_compute}

In this section, we show how the gradient of the forward model appears through
the differentiation of the joint probability density $J(\bomega)$ of \eqref{J}.
Recall (see beginning of \sref{sec:metho}) that $\bomega = (\bxi, \btheta)$,
where $\bxi$ are the parameters of the forward model $\bff(\bxi)\in\R^m$, and
$\btheta$ the parameters of the likelihood function of \eqref{likelihood}.
Therefore, let us write:
\begin{eqnarray}
J(\bomega) = J(\bxi, \btheta) &=& L(\by, \bff(\bxi), \btheta) + P_{\xi}(\bxi) + P_\theta(\btheta),\\
L(\by, \bff(\bxi), \btheta) &=& \log p(\by | \bff(\bxi), \btheta),\\
P_{\xi}(\bxi) &=& \log p(\bxi),\\
P_\theta(\btheta) &=& \log p(\btheta).
\end{eqnarray}
Using the chain rule, we have:
\begin{eqnarray}
\frac{\partial J}{\partial \xi_j} &=& \sum_{s=1}^{d_y}
\frac{\partial L}{\partial f_s}\frac{\partial f_s}{\partial \xi_j} +
\frac{\partial P_{\xi}}{\partial \xi_j},\\
\frac{\partial J}{\partial \theta_j} &=& \frac{\partial L}{\partial \theta_j}
+ \frac{\partial P_\theta}{\partial \theta_j},\\
\frac{\partial^2 J}{\partial \xi_j\partial \xi_j} &=&
\sum_{s,r=1}^{d_y} \frac{\partial^2 L}{\partial f_r\partial f_s}
\frac{\partial f_r}{\partial \xi_k}\frac{\partial f_s}{\partial \xi_j}
+ \sum_{s=1}^{d_y} \frac{\partial L}{\partial f_s}\frac{\partial^2 f_s}{\partial \xi_j\partial \xi_k}
+ \frac{\partial^2 P_{\xi}}{\partial \xi_j\partial \xi_k},\\
\frac{\partial^2 J}{\partial \theta_i\partial \theta_j} &=&
\frac{\partial^2 L}{\partial \theta_i\partial \theta_j}
+ \frac{\partial^2 P_{\xi}}{\partial \theta_i \partial \theta_j},\\
\frac{\partial^2J}{\partial \xi_j \partial \theta_k} &=&
\sum_{s=1}^{d_y}\frac{\partial^2 L}{\partial \theta_k \partial f_s}\frac{\partial f_s}{\partial \xi_j}.
\end{eqnarray}
Therefore, the Jacobian and the Hessian of the forward model are required.
However, as is obvious by close inspection of Eqs.~(\ref{eqn:L_2_c_1}),~(\ref{eqn:L_2_c_2}), and~(\ref{eqn:L_2_c_3}),
if the covariance matrices of the mixture $q(\bomega)$ of \eqref{mixture} are
diagonal, then only the diagonal elements of the Hessian of the forward model
are essential.
This is the approach we follow in our numerical examples.

\subsection{Isotropic Gaussian likelihood}
\label{ap:iso_gauss_like}
In both our numerical examples, we use the isotropic Gaussian likelihood
defined in \eqref{iso_gauss_like}.
Its logarithm is:
\begin{equation}
\label{eq:iso_likelihood}
L(\by, \bff(\bxi), \theta) = \log \calN\left(\by | \bff(\bxi), e^{2\theta}\bI\right).
\end{equation}
The required gradients are:
\begin{eqnarray*}
\frac{\partial L}{\partial f_r} &=& e^{-2\theta}\left(y_r - f_r(\bxi)\right),\\
\frac{\partial L}{\partial \theta} &=& e^{-\theta}\left(
\parallel \by - \bff(\bxi) \parallel_2^2e^{-2\theta} - d_{\xi}
\right),\\
\frac{\partial^2 L}{\partial\theta^2} &=&
e^{-\theta}\left(d_{\xi} - 3\parallel \by - \bff(\bxi) \parallel_2^2e^{-2\theta}  \right),\\
\frac{\partial^2 L}{\partial f_r\partial f_s} &=& -e^{-2\theta},\\
\frac{\partial^2 L}{\partial \theta\partial f_r} &=&
-2e^{-3\theta}\left(y_r - f_r(\bxi)\right).
\end{eqnarray*}

\subsection{Derivatives of a dynamical system}
\label{ap:dyn_adjoint}

Assume that $\bu(t, \bxi)\in\R^{d_u}$ is the solution of the following initial value problem:
\begin{equation}
\label{eqn:original_dyn}
\begin{array}{cc}
\dot{\bu} &= \bg(\bu, t, \bxi),\\
\bu(0) &= \bu_0(\bxi),
\end{array}
\end{equation}
where $\bxi\in\R^{d_{\xi}}$ are parameters.
Using the chain rule, one can show that the derivatives of $\bu(t,\bxi)$ with
respect to $\bxi$, $v_{ij} = \frac{\partial u_i}{\partial \xi_j}$, satisfy the following initial value problem:
\begin{equation}
\label{eqn:grad_dyn}
\begin{array}{cc}
\dot{v}_{ij} &= \sum_{r=1}^{d_u}\frac{\partial g_i}{\partial u_r}v_{rj} + \frac{\partial g_i}{\partial \xi_j},\\
v_{ij}(0) &= \frac{\partial u_0}{\partial \xi_j}.
\end{array}
\end{equation}
The second derivatives $w_{ijk} = \frac{\partial^2 u_i}{\partial \xi_j\partial \xi_k}$, satisfy the following initial value problem:
\begin{equation}
\label{eqn:hess_dyn}
\begin{array}{cc}
\dot{w}_{ijk} &= \sum_{r=1}^{d_u}\frac{\partial g_i}{\partial u_r}w_{rjk}
+ \sum_{r,s=1}^{d_u}\frac{\partial^2 g}{\partial u_r\partial u_s}v_{rj}v_{sk}
+ \frac{\partial^2 g_i}{\partial \xi_j\partial \xi_k},\\
w_{ijk}(0) &= \frac{\partial^2 u_0}{\partial \xi_j\partial \xi_k}.
\end{array}
\end{equation}

The numerical strategy for solving these systems is quite simple.
First, we solve \eqref{original_dyn} using an explicit runge-kutta method of
order (4)5.
Then, we use the solution as forcing in \eqref{grad_dyn} to find the gradient.
Finally, both the solution and the gradient are used as forcing in
\eqref{hess_dyn}.

\subsection{Derivatives of the diffusion equation}
\label{ap:adjoint_diffusion}
Let $u(t, \bss, \bxi) \in \R^{d_u}$  be the solution of the partial
differential equation \eqref{transport}.
The derivatives $v_{i} =
\frac{\partial u}{\partial \xi_i}$ satisfy the partial
differential equation
\begin{equation}
\label{eqn:adjoint_diff_1}
\frac{\partial v_i}{\partial t} = \nabla^2 v_i +
\frac{\partial g(t, \bx, \bxi)}{\partial \xi_i}.
\end{equation}
Similarly the second derivatives $w_{ij} = \frac{\partial^2 u}{\partial
  \xi_i \partial \xi_j}$ satisfy
\begin{equation}
\label{eqn:adjoint_diff_2}
\frac{\partial w_{ij}}{\partial t} = \nabla^2 w_{ij} +
\frac{\partial^2 g(t, \bx, \bxi)}{\partial \xi_i \partial \xi_j}.
\end{equation}
Equations
(\ref{eqn:transport}),~(\ref{eqn:adjoint_diff_1}),
~and~(\ref{eqn:adjoint_diff_2})
are solved numerically using the same space-
and time-discretization and the finite volume solver provided by
FiPy \cite{fipy}.
The only thing that changes is the source term.

\section*{References}
\bibliographystyle{plain}
\bibliography{references}

\begin{thebibliography}{10}

\bibitem{Atchade2006}
Yves~F. Atchad\'{e}.
\newblock {An adaptive version for the metropolis adjusted Langevin algorithm
  with a truncated drift}.
\newblock {\em Methodology and Computing in Applied Probability}, 8:235--254,
  2006.

\bibitem{bilionis2012}
I.~Bilionis and P.S. Koutsourelakis.
\newblock {Free energy computations by minimization of Kullback--Leibler
  divergence: An efficient adaptive biasing potential method for sparse
  representations}.
\newblock {\em Journal of Computational Physics}, 231(9):3849--3870, 2012.

\bibitem{bilionis2014solution}
I~Bilionis and N~Zabaras.
\newblock Solution of inverse problems with limited forward solver evaluations:
  a bayesian perspective.
\newblock {\em Inverse Problems}, 30(1):015004, 2014.

\bibitem{bilionisCrop2014}
Ilias Bilionis, Beth~A. Drewniak, and Emil~M. Constantinescu.
\newblock {Crop physiology calibration in CLM}.
\newblock {\em Geoscientific Model Development (under review)}, 2014.

\bibitem{bilionis2013a}
Ilias Bilionis and Nicholas Zabaras.
\newblock A stochastic optimization approach to coarse-graining using a
  relative-entropy framework.
\newblock {\em The Journal of Chemical Physics}, 138(4):--, 2013.

\bibitem{Byrd1995}
Richard~H. Byrd, Peihuang Lu, Jorge Nocedal, and Ciyou Zhu.
\newblock {A Limited Memory Algorithm for Bound Constrained Optimization},
  1995.

\bibitem{Peng2014}
Peng Chen, Nicholas Zabaras, and Ilias Bilionis.
\newblock {Uncertainty Propagation using Infinite Mixture of Gaussian Processes
  and Variational Bayesian Inference, in press}.
\newblock {\em Journal of Computational Physics}, 2014.

\bibitem{fichtner2010full}
A.~Fichtner.
\newblock {\em Full Seismic Waveform Modelling and Inversion}.
\newblock Advances in Geophysical and Environmental Mechanics and Mathematics.
  Springer, 2010.

\bibitem{Fichtner2011}
Andreas Fichtner.
\newblock {Full Seismic Waveform Modelling and Inversion}, 2011.

\bibitem{Fox2011}
Charles~W. Fox and Stephen~J. Roberts.
\newblock {A tutorial on variational Bayesian inference}.
\newblock {\em Artificial Intelligence Review}, 38(2):85--95, June 2011.

\bibitem{gershman2012}
Samuel Gershman, Matthew~D. Hoffman, and David~M. Blei.
\newblock Nonparametric variational inference.
\newblock {\em CoRR}, abs/1206.4665, 2012.

\bibitem{griewank2008evaluating}
A.~Griewank and A.~Walther.
\newblock {\em Evaluating Derivatives: Principles and Techniques of Algorithmic
  Differentiation, Second Edition}.
\newblock SIAM e-books. Society for Industrial and Applied Mathematics (SIAM,
  3600 Market Street, Floor 6, Philadelphia, PA 19104), 2008.

\bibitem{fipy}
J.E. Guyer, D.~Wheeler, and J.A. Warren.
\newblock Fipy: Partial differential equations with python.
\newblock {\em Computing in Science and Engineering}, 11(3):6--15, 2009.

\bibitem{hastings1970montecarlo}
W.~K. Hastings.
\newblock {Monte Carlo sampling methods using Markov chains and their
  applications}.
\newblock {\em Biometrika}, 57(1):97--109, 1970.

\bibitem{mary2006effective}
Mary~C. Hill and Claire~R. Tiedeman.
\newblock {\em Effective Groundwater Model Calibration: With Analysis of Data,
  Sensitivities, Predictions, and Uncertainty}.
\newblock Wiley, 2006.

\bibitem{ies_2008_huber_mfi_entropy}
Marco~F. Huber, Tim Bailey, Hugh Durrant-Whyte, and Uwe~D. Hanebeck.
\newblock On entropy approximation for gaussian mixture random vectors.
\newblock In {\em Proceedings of the 2008 IEEE International Conference on
  Multisensor Fusion and Integration for Intelligent Systems (MFI)}, Seoul,
  Südkorea, August 2008.

\bibitem{jaynes2003probability}
E.T. Jaynes and G.L. Bretthorst.
\newblock {\em Probability Theory: The Logic of Science}.
\newblock Cambridge University Press, 2003.

\bibitem{citeulike:8803391}
Eugenia Kalnay.
\newblock {\em {Atmospheric Modeling, Data Assimilation and Predictability}}.
\newblock Cambridge University Press, 1 edition, December 2002.

\bibitem{katsounaros}
I~Katsounaros, M~Dortsiou, C~Polatides, S~Preston, T~Kypraios, and G~Kyriacou.
\newblock Reaction pathways in the electrochemical reduction of nitrate on tin.
\newblock {\em Electrochimica Acta}, 71:270--276, 2012.

\bibitem{Kennedy_2001d}
M.C. Kennedy and A.~O'Hagan.
\newblock Bayesian calibration of computer models.
\newblock {\em Journal of the Royal Statistical Society: Series B (Statistical
  Methodology)}, pages 425--464, 2001.

\bibitem{citeulike:1649749}
S.~Kullback and R.~A. Leibler.
\newblock {On Information and Sufficiency}.
\newblock {\em The Annals of Mathematical Statistics}, 22(1):79--86, 1951.

\bibitem{citeulike:12630613}
Youssef Marzouk and Dongbin Xiu.
\newblock {A Stochastic Collocation Approach to Bayesian Inference in Inverse
  Problems}.
\newblock {\em Communications in Computational Physics}, 6(4):826--847, 2009.

\bibitem{McLachlan2004}
Geoffrey McLachlan and David Peel.
\newblock {\em {Finite Mixture Models}}.
\newblock John Wiley \& Sons, 2004.

\bibitem{annealing2}
N.~Metropolis, A.~W. Rosenbluth, M.~N. Rosenbluth, A.~H. Teller, and E.~Teller.
\newblock {Equation of state calculations by fast computing machines}.
\newblock {\em The Journal of Chemical Physics}, 21(6):1087--1092, 1953.

\bibitem{ionel2009}
Ionel Navon~M.
\newblock Data assimilation for numerical weather prediction: A review.
\newblock In SeonK. Park and Liang Xu, editors, {\em Data Assimilation for
  Atmospheric, Oceanic and Hydrologic Applications}, pages 21--65. Springer
  Berlin Heidelberg, 2009.

\bibitem{Plessix2006}
R.-E. Plessix.
\newblock {A review of the adjoint-state method for computing the gradient of a
  functional with geophysical applications}.
\newblock {\em Geophysical Journal International}, 167(2):495--503, November
  2006.

\bibitem{robbins1951}
Herbert Robbins and Sutton Monro.
\newblock A stochastic approximation method.
\newblock {\em The Annals of Mathematical Statistics}, 22(3):400--407, 09 1951.

\bibitem{tarantola2005inverse}
A.~Tarantola.
\newblock {\em Inverse Problem Theory and Methods for Model Parameter
  Estimation}.
\newblock Society for Industrial and Applied Mathematics (SIAM, 3600 Market
  Street, Floor 6, Philadelphia, PA 19104), 2005.

\end{thebibliography}

\end{document}